\documentclass[conference]{IEEEtran}
\IEEEoverridecommandlockouts
\usepackage{cite}
\usepackage{amsmath,amssymb,amsfonts}
\usepackage{algorithmic}
\usepackage{graphicx}
\usepackage{textcomp}
\usepackage{xcolor}
\def\BibTeX{{\rm B\kern-.05em{\sc i\kern-.025em b}\kern-.08em
    T\kern-.1667em\lower.7ex\hbox{E}\kern-.125emX}}
\usepackage{hyperref}
\usepackage{orcidlink}
\begin{document}

\title{Achieving 3D Attention via Triplet Squeeze and Excitation Block}

\author{
\IEEEauthorblockN{1\textsuperscript{st} Maan Alhazmi}
\IEEEauthorblockA{
\textit{School of Computer Science} \\
\textit{University of Leeds}\\
Leeds, UK. \\
scmaalh@leeds.ac.uk\,\orcidlink{0009-0002-4904-5560}
}
\and
\IEEEauthorblockN{2\textsuperscript{nd} Abdulrahman Altahhan}
\IEEEauthorblockA{
\textit{School of Computer Science} \\
\textit{University of Leeds}\\
Leeds, UK. \\
a.altahhan@leeds.ac.uk\,\orcidlink{0000-0003-1133-7744}
}
}

\maketitle

\begin{abstract}
The emergence of ConvNeXt and its variants has reaffirmed the conceptual and structural suitability of CNN-based models for vision tasks, re-establishing them as key players in image classification in general, and in facial expression recognition (FER) in particular. In this paper, we propose a new set of models that build on these advancements by incorporating a new set of attention mechanisms that combines Triplet attention with Squeeze-and-Excitation (TripSE) in four different variants.  We demonstrate the effectiveness of these variants by applying them to the ResNet18, DenseNet and ConvNext architectures to validate their versatility and impact. Our study shows that incorporating a TripSE block in these CNN models boosts their performances, particularly for the ConvNeXt architecture, indicating its utility. We evaluate the proposed mechanisms and associated models across four datasets, namely CIFAR100, ImageNet, FER2013 and AffectNet datasets, where ConvNext with TripSE achieves state-of-the-art results with an accuracy of \textbf{78.27\%} on the popular FER2013 dataset, a new feet for this dataset.  
\end{abstract}


\section{Introduction}
\label{sec:intro}
Advancements in CNN-based architectures have driven significant progress in various computer vision tasks \cite{LeNet, AlexNet, GoogLeNet, VGG, ResNet, MobileNetV2, EfficientNet-B0, DenseNet}, owing to their exceptional ability to extract rich, discriminative features and capture intricate patterns within visual data. Since the introduction of convolutional operations in LeNet \cite{LeNet}, CNNs have served as foundational components in subsequent vision models such as AlexNet \cite{AlexNet}, GoogleNet \cite{GoogLeNet}, VGG \cite{VGG}, and ResNet \cite{ResNet}. During that period, research efforts were primarily directed toward developing deeper neural networks, aiming to improve feature extraction capabilities and enhance model performance. The motivation behind this was to enable networks to learn more complex hierarchical representations, leading to significant breakthroughs in tasks such as image classification, object detection, and facial expression recognition.

Different studies \cite{SENet, CBAM, BAM, TripletAttention} have explored attention mechanisms to further enhance CNN-based models' performance by refining feature representations, shifting the focus from merely increasing model depth to improving the quality of learned features. Instead of relying solely on deeper architectures, attention mechanisms enable networks to selectively emphasize important regions or channels in an image, thereby improving their ability to capture meaningful patterns and dependencies. This paradigm shift has led to more efficient and interpretable models, achieving competitive performance with reduced computational complexity.

During this period, the introduction of transformers revolutionised natural language processing with their self-attention mechanism, marking a significant advancement in capturing long-range dependencies in data\cite{transformers}. Building on this success, researchers began exploring the application of transformers in other domains such as computer vision. 

Despite attempts to build transformer-based models for computer vision such as ViT \cite{transformers2} and DeiT\cite{DeiT}, the introduction of more attentive CNN-based models such as EfficientNet \cite{EfficientNet-B0} outperformed transformer-based models on widely used computer vision datasets, such as CIFAR100. Building on the same trend, ConvNeXt \cite{convnext} reestablished CNN-based models at the forefront of computer vision by proposing enhancements that surpassed the performance of transformer-based counterparts. This sparked renewed interest in utilising ConvNeXt across various computer vision tasks. 

In the domain of facial expression recognition, EmoNeXt\cite{emonext} adapted the ConvNeXt model by incorporating a Spatial Transformer Network (STN)\cite{STN} to focus on salient facial regions and used Squeeze-and-Excitation (SE)\cite{SENet} blocks to capture inter-channel dependencies. This architecture achieved a competitive performance with the state-of-the-art on the FER2013 dataset, underlining its potential for advancing facial expression recognition tasks.

In this paper, we propose a new set of models that build on these advancements by incorporating a new set of attention mechanisms that combines Triplet attention with Squeeze-and-Excitation (TripSE) in four different ways without compromising model efficiency or limiting the benefits of transfer learning.  We demonstrate the effectiveness of the TripSE block by applying it to the ResNet18, DenseNet and ConvNext architectures to validate its versatility and impact. Our study shows that incorporating the TripSE block in these CNN models boosts their performances, particularly for the ConvNeXt architecture, revealing its advantage. We evaluate the proposed mechanisms and associated models across four datasets, namely CIFAR100, ImageNet, FER2013 and AffectNet datasets, where ConvNext with TripSE achieves state-of-the-art results with an accuracy of \textbf{78.27\%} on the popular FER2013 dataset.

\section{Related Work}
In this section we review attention mechanisms in computer vision, then we review recent advancements in facial expression recognition.

\subsection{Attention Mechanisms in Computer Vision }
Attention mechanisms (AMs) have become crucial components in state-of-the-art vision models due to their performance-enhancing capabilities and potential for improved interpretability \cite{AMSurvey1}. 
Attention mechanisms enable models to focus on the most relevant information by learning to weigh the importance of different feature maps as information is passed between layers/blocks.
This is typically achieved by adding an attention layer or block between existing layers or blocks, allowing selective emphasis to be placed on crucial manifolds. A feed-forward fully connected block or a convolutional block is typically used to compute these attention weights, often employing an alignment function like a scaled dot product, followed by a softmax activation to generate scaling attention scores.

Given their stronger performance and applicability in computer vision compared to transformer-based attention (e.g., ViT), this paper investigates and proposes a range of CNN-based attention mechanisms.
Within the context of computer vision, attention mechanisms can be categorised into channel attention, spatial attention, temporal attention, and branch attention, with additional overlapping categories such as channel-spatial attention and spatial-temporal attention \cite{Attention-on-Vision-Survay}. Channel attention focuses on selecting the most relevant feature channels, determining "what to pay attention to". Examples of these techniques include Squeeze and Exitaiton networks (SE) \cite{SENet}, GSoP \cite{GSoP-Net}, SRM \cite{SRM}, GCT \cite{GCT}, ECA \cite{ECAnet}, and FCA \cite{FcaNet}. We cover SE in more detail in the Methodology section.
Spatial attention, on the other hand, emphasises important regions within an image or channel, answering "where to pay attention". Examples of these techniques include RAM \cite{RAM}, Soft-Attention and Hard-Attention \cite{ShowAttendTell}, STN \cite{STN}, and Self-Attention \cite{nonLocalNN}.
Another category of AMs focuses on the dimensionality of the passed information. For example, Triplet Attention (TA) \cite{TripletAttention} tries to answer the "where to pay attention" question by rotating the corresponding data tensor to achieve cross-dimension attention.  We cover TA in more detail in the Methodology section.
Finally, channel-spatial attention combines spatial and channel attentions to answer both the where and what questions. Examples include Residual attention network (RAN) \cite{RAN}, Convolutional Block Attention Module (CBAM) \cite{CBAM}, and Bottleneck Attention Module (BAM) \cite{BAM}.
Overall, attention mechanisms have played a transformative role in deep learning, enhancing interpretability, efficiency, and performance across computer vision tasks.

\subsection{Models and Attention Mechanism for FER}
Facial expression recognition is a challenging domain in computer vision.  Several recent studies have modified existing architectures to better accommodate facial expression recognition, often incorporating attention-based techniques. The impact of different attention mechanisms on traditional CNN models such as ResNet and MobileNetV2, have been studied in \cite{ELU+}.  The study shows that these traditional models are improved when combined with different attention mechanisms including CBAM, BAM, and TA, where the latter performed the best. \cite{residual_masking_network} introduced the Residual Masking Network which utilises a masking approach with a localisation network based on Unet to improve input feature maps, emphasising specific regions. Each Masking Block, a modified version of the Unet network, enables the Residual Masking Network to attend to important spatial features, resulting in more accurate facial expression classification. 

ConvNeXt \cite{convnext} reasserted the superiority of CNN-based models in computer vision. 
It introduces several improvements to the standard ResNet architecture. ConvNeXt features densely connected convolutional blocks within each layer, enhancing information flow and feature reuse. It also integrates multi-scale feature fusion through parallel pathways, allowing the network to capture a wide range of contextual information effectively. Additionally, ConvNeXt uses group convolutions to balance computational efficiency and expressive power, achieving top performance on various computer vision tasks while maintaining scalability.
The ConvNeXt block includes larger kernel-sized and depthwise convolutions and employs an inverted bottleneck design to reduce overall network floating-point operations (FLOPs) while improving performance. Furthermore, ConvNeXt replaces ReLU with GELU as the activation function and swaps BatchNorm (BN) for Layer Normalisation (LN), resulting in slightly better performance.

\cite{emonext} proposed the EmoNeXt architecture that enhances the ConvNeXt. Their model incorporates a Spatial Transformer Network (STN) at the beginning of the architecture. The STN allows the model to handle variations in scale, rotation, and translation by learning and applying spatial transformations to facial images. After the STN processes the inputs, they pass through several ConvNeXt blocks each followed by an SE block to enhance the extraction of salient features of these ConvNeXt blocks.

\cite{VGG_segmentation} proposed a segmentation-based VGG architecture, leveraging the widely used VGG neural network. To retain crucial feature information, a segmentation block is added before each pooling layer. These blocks perform segmentation operations to extract and highlight key features from the feature map, producing a segmentation map. To ensure the segmentation block does not miss essential features an attention mechanism is added by combining the segmentation map with the input feature map through element-wise multiplication and addition. 

\section{Methodology}

In this section, we propose a new attention block that combines the benefits of TA and SE in several novel ways, we call it TripSE block. We start by discussing TA and SE and then show several ways to combine them to achieve cross-dimensional attention that answers the "where to pay attention" and the "what to pay attention to" questions.

\subsection{Triplet Attention (TA)}
As mentioned before, different research \cite{SENet, CBAM, BAM, TripletAttention} have explored  CNN-based attention mechanisms by learning to attend to effective features for the given task. Convolutional Block Attention Module (CBAM) and Bottleneck Attention Module (BAM) compute the spatial and channel attention independently, which might result in the loss of discriminative information across other dimensions \cite{CBAM, BAM}. \cite{TripletAttention} proposed the triplet attention (TA) block, which considers cross-dimension interactions. The TA block has three branches each corresponding to a dimension $(C \times W \times H)$. 
Each branch captures the interaction of two dimensions across the third. I.e the $H$ branch captures the interactions of the different $(C \times W)$ tensors,  the $W$ branch captures the interactions of the different $(C \times H)$ tensors,  and the $C$ branch captures the interactions of the different $(H \times W)$ tensors (feature maps).
Given a tensor $x$ of shape $(C \times W \times H)$  that represents the input for the TA block, the $W$ branch rotate $x$ first to $x'$ of shape $(W \times H \times C)$, then transforms it to $(2 \times H \times C)$ by computing and concatenating the average-pooling and the max-pooling across the $W$ dimension (Z-pooling). Finally, the results are convolved with batch normalisation resulting in an output of the shape $(1 \times H \times C)$ that is passed to a sigmoid function to produce the attention weights. The $(1 \times H \times C)$ attention weights are then used to scale  $x'$ which is rotated back to $(C \times W \times H)$. The same process is followed for the $H$ and $C$ branches, except no rotation is needed for the $C$ branch. The final result is computed by averaging the results of the three branches. Although TA operates on all dimensions, each branch is essentially a separate attention mechanism that operates on 2D tensors, hence it is deemed as a 2D-based attention mechanism.

\subsection{Squeeze-and-Excitation (SE)}
To improve the representational power of convolutional neural networks, the SENet model \cite{SENet} proposed a channel attention mechanism based on the Squeeze and Excitation (SE) block. SE consists of two parts: the squeeze module that captures the global information of the input feature maps using global average pooling and the excitation module that captures the inter-channel relationships by ReLU and sigmoid activations of two FC layers. The two fully connected layers have input and output sizes corresponding to the number of channels, with a certain number of neurons in the middle that corresponds to a desired reduction ratio. Therefore, SE produces an attention vector of size C with one component for each channel that represents its importance relative to other channels. Hence, it is deemed as a 1D-based attention mechanism. The resultant attention vector is used to scale the feature maps of each channel. 

SE has two limitations. The global average pooling in the squeeze module is too simple to capture complex information, while the fully connected layers in the excitation module increase the model's complexity. Nevertheless, although SE adds some computational cost, it effectively improves channel-wise feature discrimination.
\begin{figure}[t]
\centering
\includegraphics[width=0.45\textwidth]{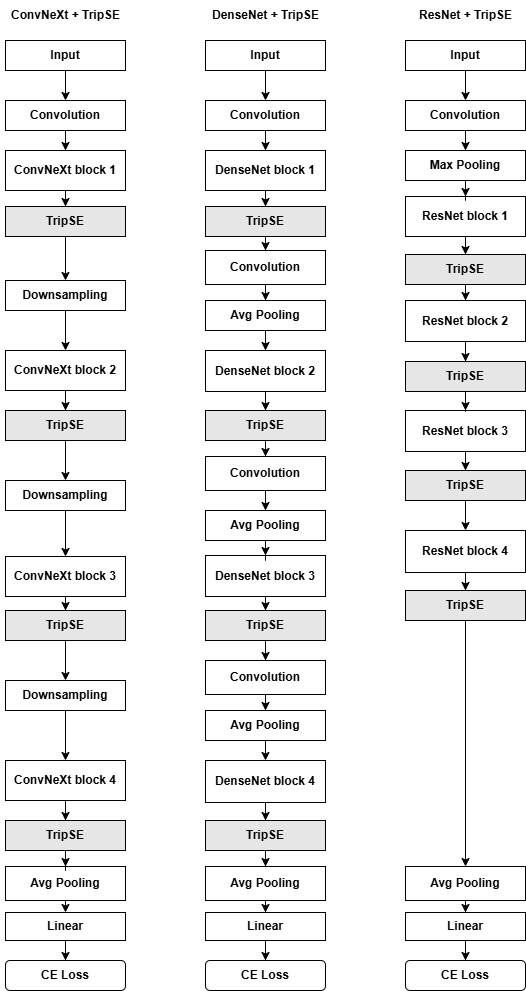}
\caption{ConvNext, DenseNet, and ResNet architectures incorporated with the TripSE block.}\label{fig:archs}
\end{figure}

\subsection{Triplet Squeeze-and-Excitation Attention (TripSE)}
As discussed in the TA subsection, the TA module employs a permutation operation to enable attention across all spatial dimensions (width, height, and channel). This permutation effectively rotates the tensor dimensions, so that information encoded within the width and height dimensions can be attended to in a channel-wise manner. We refer to the resulting channels as \textbf{rotational channels} (i.e., the original channel, width, and height dimensions after permutation).  Furthermore, the feature maps produced along the different rotational channels (including the width and height channels) are termed \textbf{rotational maps}.

Triplet Attention (TA) is a 2D attention mechanism that captures relationships between three interacting branches that act along the different rotational channels (channels, width and height dimensions) with the addition of batch normalisation. It typically involves operations like feature maps pooling and convolutions along the different rotational channels to capture their inter-dimensional or inter-rotational-channels relationships. Squeeze-and-Excitation (SE) blocks, on the other hand, are 1D-based and focus on channel-wise scaling using global pooling and fully connected layers. While TA effectively captures inter-dimensional relationships, particularly spatial and channel interactions, it might not explicitly weigh the importance of different channels in a globally informed manner. Conversely, SE excels at channel weighting but does not explicitly model the higher-order interactions captured by TA. To address these limitations, we propose a novel attention mechanism that combines the strengths of both TA and SE, aiming to enhance feature discrimination by capturing both rich inter-dimensional relationships and globally informed channel importance.

Our proposed method operates as follows: First, the input feature map is processed by a TA module. This module typically involves operations like convolutions and pooling across different dimensions to generate attention maps that capture relationships between the spatial, channel, and instance dimensions. The output of the TA module is a set of 2D attention maps, one for each dimension/branch. Concurrently, to integrate the inter-dimensional relationships captured by TA with the global channel weights from the SE block, the branch tensor undertakes an SE operation (at some stage, depending on the variant of the proposed operator -more on that in the next section), generating a rotational channel attention vector of size \textit{C}, where \textit{C} is the number of rotational channels. This 1D vector represents the global importance of each rotational channel. This vector is then used to expand and scale (via multiplicative or translational broadcasting) the branch 2D attention map to create a 3D attention map that matches the dimensions of the input tensor. The 3D attention map is then used to scale the original branch input via an element-wise multiplication. 

This translational expansion followed by element-wise multiplication allows the SE's global channel attention weights to modulate the inter-dimensional relationships learned by the TA module, effectively scaling the attention maps based on global channel importance.  This kind of affine transformation  (shifting and scaling operations) provides solid grounds for well-rounded attention that is capable of capturing different possible patterns. The resulting attention tensor represents a comprehensive representation of both rich inter-dimensional attention and globally informed channel weighting. This kind of 3D attention capability culminated in a specific variant of a set of proposed attention block variants. As we shall see later, our experiences show that such a combination of biasing and linear transformation provides maximum benefits for an attention operator.

This combined attention mechanism allows the network to focus on the most relevant spatial locations and channel interactions within the most important channels, leading to improved feature representation and potentially better performance on downstream tasks. By integrating both inter-dimensional attention and global channel attention in a multiplicative manner, our method provides a more nuanced and effective way to attend to relevant features compared to using TA or SE in isolation. 

Finally, we consolidate the resultant 3 3D rotational attention maps via either summation or further SE block to further focus the final tensor back on the usual channels, which carry more global information than their rotational counterparts. This operation also limits the loss of information caused by the averaging process.

\subsection{Variants of the TripSE Block}
We propose four different variants of TripSE blocks, each with its own advantage. The variants vary in three ways: the position of the SE block within a branch, whether we add an SE at the branch unification stage, and whether we use an additive or multiplicative scaling to scale the 2D attention map with the 1D SE attention vector for each branch. Other variants are possible, but they have been omitted for brevity.

\begin{figure}[ht]
\centering
\includegraphics[width=0.45\textwidth]{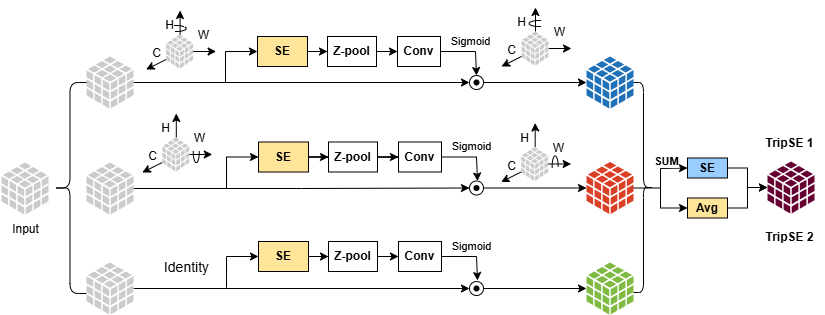}
\caption{Triplet Attention with Squeeze-and-Excitation Blocks 1 (TripSE1) and 2 (TripSE2). TripSE1 is achieved by activating the sky-blue block and deactivating the yellow blocks, where TripSE2 is achieved visa-versa.}\label{fig_ta}
\end{figure}

We start with the simplest forms of our proposed 3D combination of the TA and SE attention blocks. Figure~\ref{fig_ta} illustrates two variants of the proposed 3D attention block, TripSE1 and TripSE2. TripSE1 is achieved by activating the sky-blue block and deactivating the yellow blocks, while, in TripSE2, we inverse these activations. Hence, instead of passing the summation of the branches to one SE block at the end, we activate an SE block at the beginning of each branch to act on the permuted tensor. The channel-weighted tensor is then passed to the usual TA operations of z-pooling and convolution, followed by sigmoid.

\begin{figure}[ht]
\centering
\includegraphics[width=0.45\textwidth]{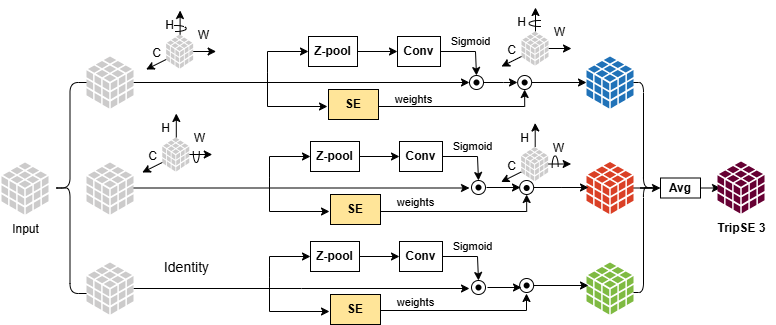}
\caption{Triplet Attention with Squeeze-and-Excitation Block 3 (TripSE 3).}\label{fig_TripSE3}
\end{figure}

In TripSE3, instead of adding the SE block at the beginning of each branch in a sequential fashion, we added the SE block in a parallel fashion, which means the branch-permuted input goes through both the original TA branch and an SE block. 
Contrary to TripSE2, the weights of the SE block will be used to scale the \textit{output} of the branch-permuted input tensor. There is no SE block at the unification of the three branches. Figure~\ref{fig_TripSE3} illustrates the TripSE3 block.

\begin{figure}[ht]

\centering
        \includegraphics[width=0.45\textwidth]{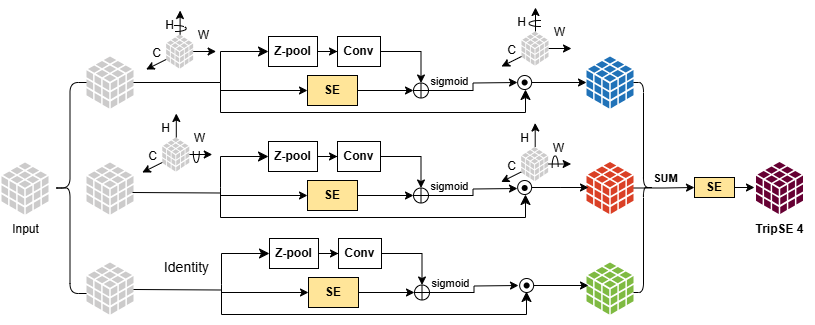}

\caption{Triplet Attention with Squeeze-and-Excitation Block 4 (TripSE  4).}\label{fig_TripSE4}
\end{figure}

In the previously explained TripSE blocks, the output of each branch is a single 2D feature map that is used to scale the branch permuted input tensor. TripSE4 represents the culmination of our proposed attention blocks, incorporating parallel SE blocks for each of the three branches, similar to TripSE3. However, each SE block in TripSE4 is enhanced by an affine transformation that utilizes a shift (translation) operation, yielding a cohesive and powerful final processing block. 

Therefore, instead of producing a single weighting plane, which only scales the input in two dimensions, we produce a weighting tensor that allows for a more nuanced, three-dimensional scaling of the permuted input tensor. This is achieved by passing the permuted tensor of each branch to the corresponding branch's SE block in parallel, as we did in TripSE3. Following that, before passing the results of the branch and the SE block to the sigmoid function, we added them element-wise and then passed the result to a sigmoid function. This produces the desired weighting tensor used to scale the permuted input tensor. We also passed the sum of the permuted input tensors from all three branches to an SE block, similar to how global channel attention was calculated in TripSE1. Figure~\ref{fig_TripSE4} illustrates the TripSE4 block.

\subsection{Incorporating the TripSE block}
We incorporated the TripSE attention block into three well-known architectures: ConvNeXt, DenseNet, and ResNet. Specifically, we added a TripSE attention block after each group of the corresponding network block. This design choice ensures minimal computational overhead, as only four attention blocks were introduced, preserving the model’s efficiency. Moreover, since we are using transfer-learning this will not disturb the layers inside the group of blocks as we did not change the inner blocks of the models. In other words, this will make the structural enhancements that we are applying to the existing architectures more compatible. The resultant models seem to be more effective in dealing with computer vision tasks, especially facial expression recognition tasks than the original ones. Fig.~\ref{fig:archs} illustrates the architectures of the ConvNeXt, DenseNet, and ResNet incorporated with TripSE.

\section{Experiments}

To investigate the effectiveness of incorporating existing architectures with the proposed TripSE operator, we chose two widely used traditional CNN models, ResNet18 and DenseNet; then we moved to the more recent CNN model, ConvNeXt. In terms of datasets, we assess our models on four datasets: CIFAR100, ImageNet, and two FER datasets, FER2013 and AffectNet.

First, we conducted an initial study on ImageNet using ResNet18 and ConvNeXt-T with the purpose of gaining insight into the effectiveness of the proposed operator. In this experiment, we trained these models with and without the incorporation of the TripSE block, without the utilisation of transfer learning (i.e. we trained from scratch).

Second, we moved to explore the incorporation of the proposed block while utilising transfer learning. In this experiment, we finetuned ResNet18, DenseNet, and ConvNeXt with and without the incorporation of the TripSE block on CIFAR100, FER2013, and AffectNet. 

Finally, to ensure a fair comparison, we finetuned the ConvNeXt model incorporated with the SE block only, and with the TA block only on the FER2013 dataset.

\subsection{Datasets}
AffectNet is a large-scale real-world dataset for facial expression recognition. It contains 287,651 training facial images expressing various emotions and 4,000
manually annotated validation set. It was collected from the internet using emotion-related keywords in six different languages. Approximately half of these images were manually annotated with both categorical emotions (such as happiness, sadness, and anger) and valence-arousal values. AffectNet is one of the largest datasets for facial emotion recognition.

The Facial Expression Recognition 2013 (FER2013), is a highly regarded benchmark dataset in the domain of facial emotion recognition. It was created to promote research in automated facial expression analysis and contains 35,887 labelled images of facial expressions, classified into seven emotional categories: anger, disgust, fear, happiness, sadness, surprise, and neutral. The dataset is split into three sections: 28,709 images for training, 3,589 images for validation, and 3,589 images for testing. All images are in grayscale and feature a diverse range of individuals of various characteristics. The FER2013 dataset was sourced from the internet and annotated by human evaluators. Researchers often use this dataset to train and assess models for facial expression recognition and generation.

\subsection{Implementation Details}
Following the work presented in \cite{residual_masking_network} \cite{VGG_segmentation}, the images of the FER2013 were resized from 48$^2$ to 224$^2$. Following that we have applied two data augmentation techniques, horizontal flip, and rotation from -30 to 30 degrees. All of the experiments were conducted using V100 GPUs with 32GB memory. A learning rate scheduler was used where we adjusted the learning rate by a factor of 10 when the model stopped improving on the validation set for three consecutive epochs. For other datasets, we followed the same procedure, with the only exception being the learning rate scheduler. Instead of the previous approach, we used a step learning rate scheduler, reducing the learning rate by a factor of 10 every 10 epochs. For the CIFAR100 and FER2013 datasets, the models were trained for 100 epochs, whereas for the AffectNet dataset, training was limited to 10 epochs. Table~\ref{impdetails} shows the implementation details.

\begin{table}[h]
\caption{Implementation details.}\label{impdetails}%
\centering
\begin{tabular}{lc}
\hline
\textbf{Hyperparameter} & \textbf{finetuning}\\
\hline
Batch size & 512 \\
Initial learning rate   & 0.001  \\
Minimum learning rate   & 1e$^{-6}$  \\
Image size   &224$^2$  \\
Optimiser   & RAdam  \\
GPUs   & V100 x 4  \\
\hline
\end{tabular}
\end{table}

\section{Results}

This section presents the results of the proposed modifications on different classification tasks.

\subsection{Initial Study}

To start exploring the proposed TripSE blocks, we trained the original tiny ConvNeXt model and the tiny ConvNeXt model incorporated with the newly proposed TripSE1 attention block from scratch on two benchmark datasets, CIFAR100 and ImageNet. We followed the same training procedure on the original ConvNeXt paper.

\begin{table}[ht]
\caption{Performance comparison between the ResNet18, DenseNet, ConvNeXt-S
models and the models incorporated with the proposed
TripSE block for both the pretrianing on ImageNet and the finetuning on CIFAR100, FER2013, and AffectNet.}\label{tab2}%
\begin{tabular}{lcccc}
\hline
\textbf{Model} & ImageNet & CIFAR100 & FER2013 & AffectNet \\

\hline
ResNet18    & 70.94\footnotemark[1] & \textbf{67.56}\footnotemark[1] & 71.11 & 60.71 \\
ResNet18 + TripSE1  & \textbf{71.74}\footnotemark[1] & 67.08\footnotemark[1] & \textbf{74.12} &  \textbf{61.11} \\

DenseNet & - & 85.33 & 74.63 & \textbf{62.16}\\
DenseNet + TripSE1 & - & \textbf{85.57}& \textbf{75.95} & 61.10\\

ConvNeXt & 82.20\footnotemark[1] & 80.39\footnotemark[1] & 77.19 & 62.87\\
ConvNeXt + TripSE1& \textbf{82.30}\footnotemark[1] & \textbf{81.53}\footnotemark[1] & \textbf{78.08} & \textbf{64.16} \\
\hline
\end{tabular}
\\
\\ \footnotesize[1]{ Trained from scrtach}
\end{table}

On CIFAR100, we achieved a performance increase of over one percent, from \textbf{80.39\%} using the original tiny ConvNeXt model to \textbf{81.53\%} using the tiny ConvNeXt model incorporated with the newly proposed TripSE attention block. Such results are promising and illustrate the potential of the proposed attention block and the proposed architectural modification. On the other hand, using ResNet18 with TripSE did not achieve any performance gain as the original ResNet18 scored \textbf{67.56\%}, while the TripSE incorporated ResNet18 scored \textbf{67.08\%}.

On ImageNet, a slight performance increase was achieved, from \textbf{82.20\%} using the original tiny ConvNeXt model to \textbf{82.30\%} using the tiny ConvNeXt model incorporated with the newly proposed TripSE attention block. Since the incorporation with the ConvNeXt model didn't achieve a significant improvement on ImageNet, we tested it with ResNet18. The results on ResNet18 show consistent observation, where an increase of almost one present was achieved from \textbf{70.94\%} to \textbf{71.74\%}. 

Moreover, the proposed modification did not harm the efficiency of the model. The original tiny ConvNeXt model has 28.6M trainable parameters, while the tiny ConvNext incorporated with the TripSE block has 28.7M trainable parameters. Such a small increase in the number of parameters is negligible considering such a performance gain. Table~\ref{tab2} illustrates the results of the tiny ConvNeXt model and the tiny ConvNeXt model incorporated with the proposed TripSE attention block on the CIFAR100 and ImageNet datasets. Besides, it shows the results of ResNet18 with and without the incorporation of the TripSE block.

\subsection{Results of Different Classification Tasks}
At this stage of our experiments, we aimed at incorporating the TripSE block with the chosen models, and train them while utilising transfer learning 

First, the pretrained ResNet18 and ResNet18 + TripSE on ImageNet were finetuned on the FER2013 dataset. The results are consistent and persistent, where a significant increase in performance of three percent was achieved using the ResNet18 + TripSE compared with the original ResNet18 from \textbf{71.11\%} to \textbf{74.12\%}. Table~\ref{tab2} illustrates the performance comparison of the ResNet18 model and ResNet18 model incorporated with the proposed TripSE attention block for both the pretrianing on ImageNet and the finetuning on FER2013.

Furthermore, we evaluated the proposed TripSE attention block integrated with DenseNet. Specifically, we finetuned a pretrained DenseNet121 model with and without incorporating the TripSE block on two of the chosen datasets: CIFAR100 and FER2013. The results consistently demonstrated the effectiveness of the TripSE block. While the performance improvement on CIFAR-100 was modest, a more substantial gain was observed on FER2013, with the model achieving an accuracy of \textbf{75.95\%}, reflecting an increase of over one percentage point.  

\subsection{Results on FER Task}
Since the TripSE block demonstrates a promising effect on the FER task, we selected a pretrained Tiny ConvNeXt model (ConvNeXt-T) and finetuned it on the FER2013 dataset. The accuracy score of the original model is \textbf{77.20\%}. Following that, we added a SE block after each group of ConvNeXt blocks and finetuned it on the FER2013. This has reduced the accuracy score to \textbf{76.93\%}. Then, we incorporated the ConvNeXt with a TA block instead of the SE block, which also reduced the accuracy score to \textbf{76.51\%}. Finally, we incorporated the ConvNeXt with the newly proposed TripSE1 block, where we achieved the state-of-the-art result, to the best of our knowledge, at \textbf{77.86\%} showing an increase of one percent compared with the current reproducible state-of-the-art at \textbf{76.82\%} \cite{residual_masking_network}.

We took a step further and tested the proposed TripSE1 block with the small-size ConvNeXt model (ConvNeXt-S) exactly as we did with the Tiny model. The Results are even more interesting. We were able to push the performance on FER2013 to \textbf{78.08\%} which is also another state-of-the-art result for this dataset. Note that the original model failed to achieve any performance gain when we moved to the larger size model maintaining similar performance at \textbf{77.19\%}, which demonstrates the effectiveness of the proposed method. Table~\ref{tabsota} compares the performance of the current state-of-the-art, the ConvNeXt models, the ConvNeXt models incorporated with the TA block, the SE block, and the proposed TripSE attention block on the FER2013.

Finally, we tested the proposed models on a large-scale FER dataset, namely AffectNet. Similar trend is observed where we achieved more than one percent increment in performance using the ConvNeXt-S + TripSE1 from \textbf{62.87\%} to \textbf{64.16\%}. On ResNet18, a slight increase in performance was observed as well from \textbf{60.71\%} to \textbf{61.11\%}. On DenseNet, however, the modified architecture didn't achieve any performance gain scoring \textbf{61.10\%} where the original model scored \textbf{62.16\%}, see table ~\ref{tab2}.

\begin{table}[ht]
\caption{Performance comparison of the current state-of-the-art, the Tiny ConvNeXt model, the Tiny ConvNeXt model incorporated with the TA block, the SE block, and the proposed TripSE attention blocks on the FER2013 dataset.}\label{tabsota}%
\setlength{\tabcolsep}{2pt}
\centering
\begin{tabular}{lcc}
\hline
\textbf{Model} & \textbf{Parsms. $\times 10^6$}&\textbf{Accuracy \%} \\
\hline
Human Accuracy \cite{FER2013} & - &65 $\pm5$   \\
Deep Emotion \cite{DeepEmotion} & - & 70.02  \\
Inception \cite{khaireddin2021facial}& 23.9&71.60  \\
Ad-Corre \cite{Ad-corre} & 26.1  &72.03  \\
SE-Net50 \cite{khanzada2020facial} & 27 & 72.50 \\
ResNet50 \cite{khanzada2020facial}& 25.6 & 73.20  \\
VGG \cite{khaireddin2021facial}& 139.5 & 73.28  \\
CBAM ResNet50 \cite{VGG19seg}& 28.5& 73.32  \\
ResNet34v2 \cite{VGG19seg}& 21.8 &73.65   \\
ResMaskingNet \cite{residual_masking_network}& 142.9 & 74.14  \\
LHC-Net \cite{LCH} & 32.4 & 74.42 \\
Segmentation VGG-19 \cite{VGG19seg} & 191.5 & 75.97 \\
RMN + 6 other CNNs \cite{residual_masking_network} & - & 76.82\\
\hline
ConvNeXt-T\footnotemark[1] & 28.6 & 77.19\\
ConvNeXt-T + TA\footnotemark[1] & 28.6 & 76.51\\
ConvNeXt-T + SE\footnotemark[1] & 28.7 & 76.93\\
ConvNeXt-T + TripSE1 (ours)\footnotemark[1] & 28.7 & 77.86\\
ConvNeXt-S\footnotemark[1] & 50.0 & 77.19\\
ConvNeXt-S + TA\footnotemark[1] & 50.0 & 76.81\\
ConvNeXt-S + SE\footnotemark[1] (r=16) \footnotemark[2]& 50.1 & 76.54\\
ConvNeXt-S + SE\footnotemark[1] (r=1) \footnotemark[2]& 50.1 & 76.66\\
ConvNeXt-S + TripSE1 (ours)\footnotemark[1] (r=16) \footnotemark[2]& 50.1 & 78.08\\
ConvNeXt-S + TripSE1 (ours)\footnotemark[1] (r=1) \footnotemark[2]& 50.1 & 76.61\\
ConvNeXt-S + TripSE2 (ours)\footnotemark[1] & 50.1 & 77.08\\
ConvNeXt-S + TripSE3 (ours)\footnotemark[1] & 50.1 &76.90\\
ConvNeXt-S + TripSE4 (ours)\footnotemark[1] (r=16) \footnotemark[2]& 50.1 & 76.71\\
ConvNeXt-S + TripSE4 (ours)\footnotemark[1] (r=1) \footnotemark[2]& 50.1 & \textbf{78.27}\\
\hline
\end{tabular}
\\ \footnotesize[1]{ Finetuned on FER2013 using the weights available in the original ConvNeXt repo at \url{https://github.com/facebookresearch/ConvNeXt}}

\footnotesize[2]{ 'r' is the reduction ratio used in the SE block.}
\end{table}

More importantly, we have conducted experiments to test the other variants of the TripSE1. Although TripSE2 and TripSE3 are more complicated and well-designed; they did not outperform TripSE1. The results of the TripSE2 and TripSE3 on FER2013 are \textbf{77.08\%} and \textbf{76.9\%}, respectively. This is due to not incorporating the 3D attention in these two nor incorporating a final SE bloc at the unification of the branches both of which are important to help the attention block model the flexible relationship of its tensor and its weights. This is also due to being incompatible with the existing pretrained models. Therefore, they must be tested in a different case study. On the other hand, TripSE4 as expected improved the performance on FER2013, achieving the best state-of-the-art results of \textbf{78.27\%} which is an unprecedented feat on the FER2013 dataset.

\section{Conclusion}

In conclusion, we have proposed novel integrations of Triplet Attention and Squeeze-and-Excitation attention, culminating in a 3D attention block capable of capturing complex attention relationships in a more nuanced manner.  By incorporating both inter-dimensional and global channel attention, our approach offers a more comprehensive and effective way to model feature dependencies. Our proposed TripSE attention block offers a promising direction for enhancing feature representation and potentially improving performance in various computer vision tasks.

Our study highlights the effectiveness of incorporating the TripSE attention mechanism into CNN-based architectures, reaffirming their strength in vision tasks, particularly in facial expression recognition. By incorporating TripSE into ResNet18, DenseNet, and ConvNeXt, we demonstrate significant performance improvements, with ConvNeXt benefiting the most. 

Our extensive evaluation across multiple datasets, including CIFAR100, ImageNet, FER2013, and AffectNet, confirms the versatility and impact of our approach. Notably, ConvNeXt with TripSE achieves a state-of-the-art accuracy of 78.27\% on FER2013, setting a new benchmark for this dataset. These findings emphasise the continued relevance of CNNs and the potential of attention mechanisms to further enhance their capabilities in image classification tasks.

\bibliographystyle{IEEEtran}
\bibliography{TripSE}

\end{document}